\documentclass[runningheads]{llncs}

\usepackage{amsmath,amsfonts,bm}

\newcommand{\defeq}{\mathrel{:\mkern-0.25mu=}}

\def\eqref#1{equation~\ref{#1}}

\def\1{\bm{1}}

\DeclareMathOperator*{\E}{\mathbb{E}}

\def\p{{\textnormal{p}}}
\def\P{{\textnormal{p}}}

\def\rx{{X}}
\def\ry{{Y}}
\def\rz{{Z}}

\DeclareMathAlphabet{\mathsfit}{\encodingdefault}{\sfdefault}{m}{sl}
\SetMathAlphabet{\mathsfit}{bold}{\encodingdefault}{\sfdefault}{bx}{n}

\usepackage{booktabs}
\usepackage{graphicx}
\usepackage{url}
\usepackage[ruled]{algorithm}      
\usepackage{algpseudocode}  
\usepackage{algorithmicx} 
\usepackage{varwidth} 
\usepackage{setspace}
\usepackage{wrapfig}
\usepackage{hyperref}

\usepackage[utf8]{inputenc}
\usepackage[T1]{fontenc}

\begin{document}
\title{Unsupervised Adversarial Image Inpainting}
%
%
\author{Arthur Pajot\inst{1} \and
Emmanuel de B\'ezenac\inst{1} \and
Patrick Gallinari\inst{1,2}}
\authorrunning{Pajot et al.}
%
\institute{Sorbonne Universit\'es, UMR 7606, LIP6, F-75005 Paris, France \and
Criteo AI Lab, Paris, France\\}
\maketitle              
\begin{abstract}
We consider inpainting in an unsupervised setting where there is neither access to paired nor  unpaired training data. The only available information is provided by the uncomplete observations and the inpainting process statistics. In this context, an observation should give rise to several plausible reconstructions which amounts at learning a distribution over the space of reconstructed images. We model the reconstruction process by using a conditional GAN with constraints on the stochastic component that introduce an explicit dependency between this component and the generated output. This allows us sampling from the latent component in order to generate a distribution of images associated to an observation. We demonstrate the capacity of our model on several image datasets: faces (CelebA), food images (Recipe-1M) and bedrooms (LSUN Bedrooms) with different types of imputation masks. The approach yields comparable performance to model variants trained with additional supervision.

\end{abstract}
%
%
%

\section{Introduction}

Image inpainting is the process of recovering missing structured information in images. This allows fill in missing parts, removing unwanted objects and it is related to other tasks such as super-resolution or denoising. This is a basic brick in many vision or image analysis applications. For example, satellite images are often incomplete (e.g. clouds) and require reconstructing the missing information. Another example is security or self driving cars vision systems, which are confronted with partial and incomplete measurements requiring sometimes the reconstruction of missing parts.
Image completion or inpainting is generally a complex inverse problem, it is usually under specified so that there is not a unique solution. Initial attempts to image completion operated on a single target image, trying to cope with missing information by using the statistics and characteristics of this image only \cite{Barnes2009,Simakov2008}. More recently, learning based approaches exploiting the information of large image corpora have been used for this problem. Most of these recent approaches exploit the structure of convolutional neural network architectures which offer an adequate prior for images \cite{deepimageprior}, together with the image sharpening ability of adversarial training. Since its publication in 2014 \cite{goodfellow2014generative}, the latter principle has given rise to an overabundant litterature on image generation, and has been used in most recent work on inpainting as well.




Despite exhibiting interesting results \cite{pathak2016context,multiscale_inpainting,attention_inpainting,yu2018free}, these methods all require some form of supervision, either observation measurement-image pairs, or unpaired samples from observations and complete images. For many real-world problems, obtaining these samples is too expensive and/or impractical. The above examples of satellite imagery and self driving car vision systems are typical cases for which no supervision with a complete signal is possible and one has to learn from incomplete and corrupted images directly. As mentionned above, image inpainting is usually an ill posed problem so that multiple signal reconstructions could explain a corrupted image. Most approaches propose a unique image recosntruction among all the possible ones. A more challenging task consists in learning the distribution of the plausible reconstructions. A common approach to distribution learning  consists in training a neural model to map a latent code taken from an easy to sample distribution, to a target output domain. By sampling the latent space, the image distribution can then be recovered. The generator performing the mapping is then supposed to learn associating latent codes with some representations of the information missing in the observation. In practice, enforcing the latent code to be taken into account by the generator so as to effectively learn the target distribution is non-trivial,  see section \ref{seq:latent_renconstruction}.

We propose here a model for the task of inpainting, which allows us to learn the distribution of reconstructed images in a completly unsupervised setting. We suppose that there is no access to any form of target distribution and that learning can only exploit incomplete and corrupted observations.
This model optimizes an  objective function combining an adversarial loss for recovering realistic and sharp image signals, and a reconstruction loss that constrains an explicit dependency between the stochastic variables and the reconstructed images thus allowing the generation of diverse reconstructions for a a same incomplete observation (see Section \ref{sec:model}). 
This model is evaluated and compared to baselines on 3 image datasets, CelebA \cite{celeba}, LSUN Bedrooms \cite{lsun}, Recipe-1M \cite{recipe}. We experiment with different types of masking processes corrupting the images.

Our contributions are :

\begin{itemize}
    \item A  framework for large-scale image inpainting learning in an fully unsupervised context.
    \item A new model for incorporating stochastic components that enforces an explicit dependency between latent variables and generated outputs. By sampling from this latent component, a distribution of reconstructed images can then be generated.
    \item Extensive evaluations on a three image datasets with different measurement processes.
\end{itemize}

\section{Model}
\label{sec:model}
\subsection{Background}


We suppose that there exists a domain of uncorrupted signals $X$ with distribution $p_\rx$, but we only have access to incomplete measurements from domain $Y$ with distribution $\P_\ry$. The masking process is  modeled through a stochastic operator $F : X  \rightarrow Y$ mapping signals $x$ to their associated observations $y$. We will refer to $F$ as the measurement process. $F$ is parameterized by mask $m \in M$ that we can sample with $\P_M$, the mask distribution. Thus, given signal $x$, we can simulate the associated observation $y$ by sampling $m$ from $\P_M$, and then calculating: 

\begin{equation}
\label{eq:measument_process}
y = F(x; m) = x \odot m  + \tau \bar{m} 
\end{equation}

Here, $m$ is a mask with the same size as $x$ and with components in $\{0,1\}$, where $0$ holds for the masked information, $\bar{m}$ denotes the complement of $m$, $\odot$ element-wise multiplication and $\tau$ the value applied on the masked region. $F$ is differentiable \textit{w.r.t.} its first argument $x$. $M$ and $\rx$ are assumed to be independent. Different distributions of $M$ will be considered (refer to section \ref{sec:experiment}). For simplification’s sake, we will assume $\tau = 0$ for the rest of the paper.

\begin{figure}[h]
\begin{center}
\includegraphics[width=0.50\textwidth]{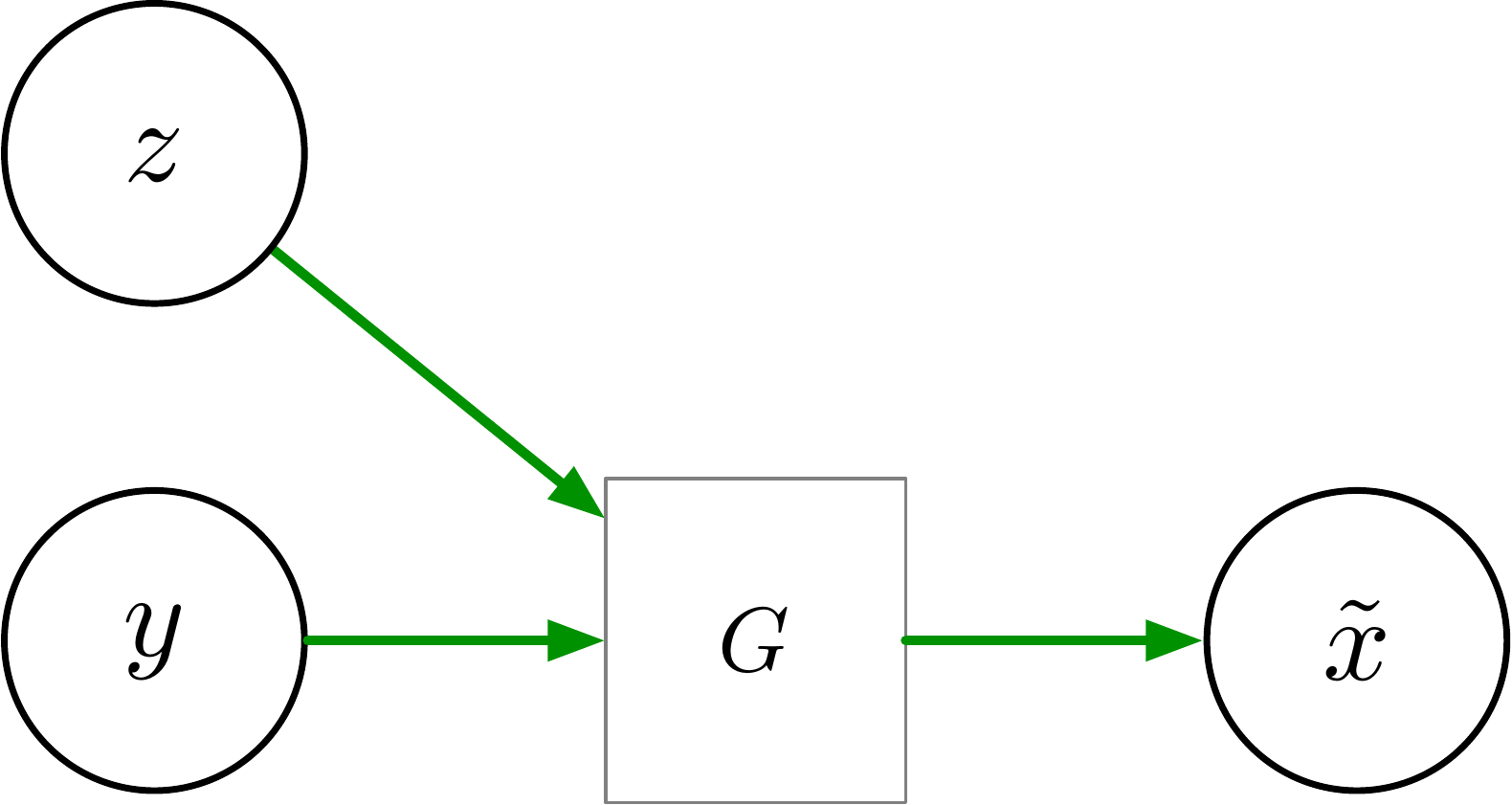}
\end{center}
   \caption{A corrupted image $y$ is sampled from $Y$ together with a latent $z$ from $Z$. The generator $G$ then produces $\tilde{x}=G(y,z)$ a reconstructed image.}
\label{fig:inference}
\end{figure}

Our goal is to learn the distribution of the semantically plausible reconstructions $x$ for an observation $y$. Said otherwise, one wants a mapping $G$ which given an observation $y$ and a random variable $z$ produces a reconstruction of a plausible underlying signal $x$, i.e. $x=G(y,z)$. Since for a given $y$ there are many possible reconstructions $x$, one wants different samples $z$ to lead to different realizations $x$. The learned mapping $G$ will associate to a given sample $z$, some information not contained in the observation $y$ for generating a full image $x$. For example if the generator is presented with a picture of a face with missing eyes, $G$ could learn to associate to different $z$ values, different representations of the color and shape of the eyes, thus generating diverse reconstructions $x$ for the same observation $y$. We will adopt throughout this work a conditional adversarial approach, and $G$ will then act as a generator. The model used for inference is illustrated in Figure \ref{fig:inference}.\\


\subsection{Adversarial loss}
\label{sec:adversarial}

Since we do not have access to the true images in $X$, but only to data sampled from partial observations in $Y$ and to the measurement process $F$, a natural adversarial formulation of the problem is the following:

\begin{equation}
    \mathcal{L}^{GAN}(G, D) \defeq
    \E_{\substack{y \sim \P_\ry}} \big[ \log D(y) \big] 
    + \E_{\substack{y \sim \P_\ry \\ m \sim \P_m \\ z \sim \P_z}} \big[ \log\big(1 - D(F[G(y, z), m]\big) \big]
    \label{eq:adversarial_z_x_1}
\end{equation}

\begin{figure}[h]
\begin{center}
\includegraphics[width=0.85\textwidth]{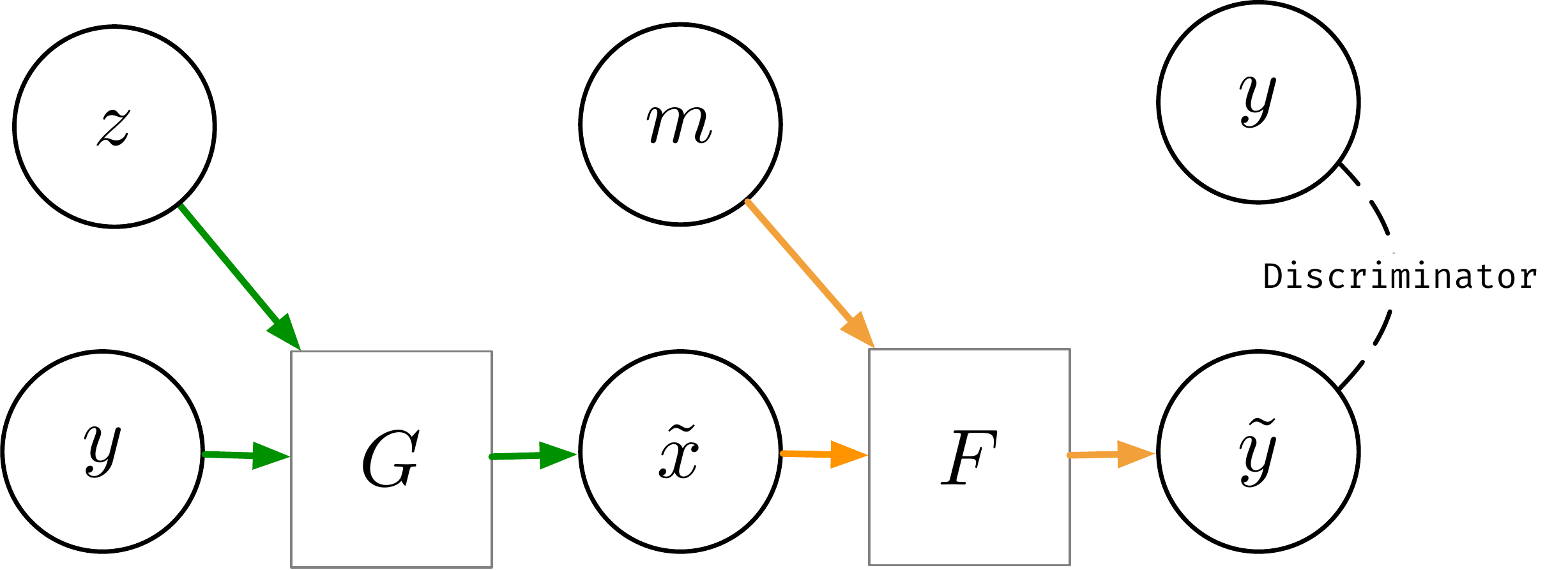}
\end{center}
   \caption{Generative model for unsupervised learning of inpainting distributions. The discriminator is trained to differentiate between the generated $\Tilde{y}=F(G(y, z), m)$ and samples $y$ from the observation dataset $Y$.}
\label{fig:gan_loss}
\end{figure}

In this equation (see also figure \ref{fig:gan_loss}), $D(y)$ is a binary discriminator trained to separate observations from generated data, $G(y, z)$ is a reconstruction $\tilde{x}$ of the underlying signal $x$, $F[G(y, z), m]$ is the signal $\tilde{y}$ obtained from $\tilde{x}$ when applying mask $m$ sampled from $p_m$. This mimics the measurement process: $\tilde{y}$ is obtained from the generated $\tilde{x}$ in the same way as $y$ is supposed to be obtained from the real $x$. Note that $m$ being a stochastic variable, the $m$ samples for $y$ and $\tilde{y}$ will presumably be different. Given this loss function, training proceeds as usual by solving:

\begin{equation}
   min_{\substack{G}} max_{\substack{D}} \mathcal{L}^{GAN}(G, D) 
   \label{minmax}
\end{equation}

Equation \ref{minmax} means that one wants to train a generator $G$ so that the fake observation $\tilde{y} = F[G(y, z), m]$ has a distribution similar to the one of the observations $y$. The difference w.r.t. classical conditional GAN formulation is that since we are here in an unsupervised setting with only partial observations. Adversarial training is then performed on observations  $y$ and on their simulated reconstructions $\tilde{y}$ instead of on complete images $x$ and $\Tilde{x}$.
Similar ideas have been used in \cite{ambientgan} and \cite{pajot_unsupervised_2019}.

This model holds for any stochastically generated noise or missing value. It could easily be refined for the inpainting task by focusing the image reconstruction process on the missing part of the image. In Equation \ref{eq:adversarial_z_x_1}, the whole image $\tilde{x}$ is supposed to be reconstructed by the generator $G$. Since one already knows $y$, there is no need to renconstruct it and one could simply plug $y$ for $x$ in the inpainting process. Let us suppose that one can retrieve the mask $m$ from the observation $y$. This is not very restrictive since in most situations, this will amount at finding the pixels of $y$ which  equals to $\tau$ (in our case $0$). The reconstruction process then becomes:

\begin{equation}
\label{eq:generator}
 \tilde{x} = G(y, z) \odot \bar{m} + y
\end{equation}

This  ensures that the observed part $y$ remains unchanged in $\tilde{x}$.
The loss in \ref{eq:adversarial_z_x_1} now becomes:

\begin{equation}
    \mathcal{L}^{GAN}(G, D) \defeq
    \E_{\substack{y \sim \P_\ry}} \big[ \log D(y) \big] 
    + \E_{\substack{y \sim \P_\ry \\ m \sim \P_m \\ z \sim \P_z}} \big[ \log\big(1 - D(F[G(y, z)+y, m]\big) \big]
    \label{eq:adversarial_z_x}
\end{equation}

Figure \ref{fig:gan_loss} illustrates the corresponding adversarial model.

\subsection{Latent Reconstruction Loss}
\label{seq:latent_renconstruction}

Although the above model has the potential for learning inpainting distributions, in practice the generator trivially learns a deterministic mapping from observations $y$ to reconstructed signals $\tilde{x}$. This has also been described in other settings, e.g. \cite{zhu_toward_2017,augmentedcyclegan}.
We then propose two complementary losses which help to enforce the dependency on the stochastic component.

\subsubsection{Encoding $z$ loss.}

A simple way to condition the generator $G$ on the stochastic component $z$ is to constrain the generator dependent outputs $\tilde{x}$ or $\tilde{y}$ to contain information from $z$. This could be implemented by training the model to recover $z$ from the output $\tilde{y}$ for example. Let us denote $\tilde{z} \defeq E(\tilde{y})$ this reconstruction with $E : Y \rightarrow Z$ a mapping to be learned. This simply amounts at adding the following term to the loss in equation \ref{eq:adversarial_z_x}:

\begin{equation}
    \mathcal{L}^{z}(G,E) \defeq \E_{\substack{z \sim \P_z \\ y \sim \P_{\ry} \\ m \sim \P_m}} \left\Vert z - E(F[G(y, z) + y), \bar{m}])\right\Vert_2^2
    \label{eq:rec_z}
\end{equation}

This loss enforces the generator to use the information generated from the latent $z$ into the reconstruction of the masked input. However, although this partially fulfills our goal, this is not sufficient, since as shown in \cite{augmentedcyclegan} and \cite{chu2017cyclegan}, this auxiliary loss exhibits a "stenography" behavior: the corresponding model tends to hide information generated by $G$ when using $z$ as stochastic input, in a visually imperceptible way in the reconstructed image $x$. One needs a more direct way of enforcing diversity in the generated images. In order to alleviate this issue, we propose a second auxiliary loss. 

\subsubsection{Encoding $y$ Loss.}

\begin{figure}
\begin{center}
\includegraphics[width=0.90\textwidth]{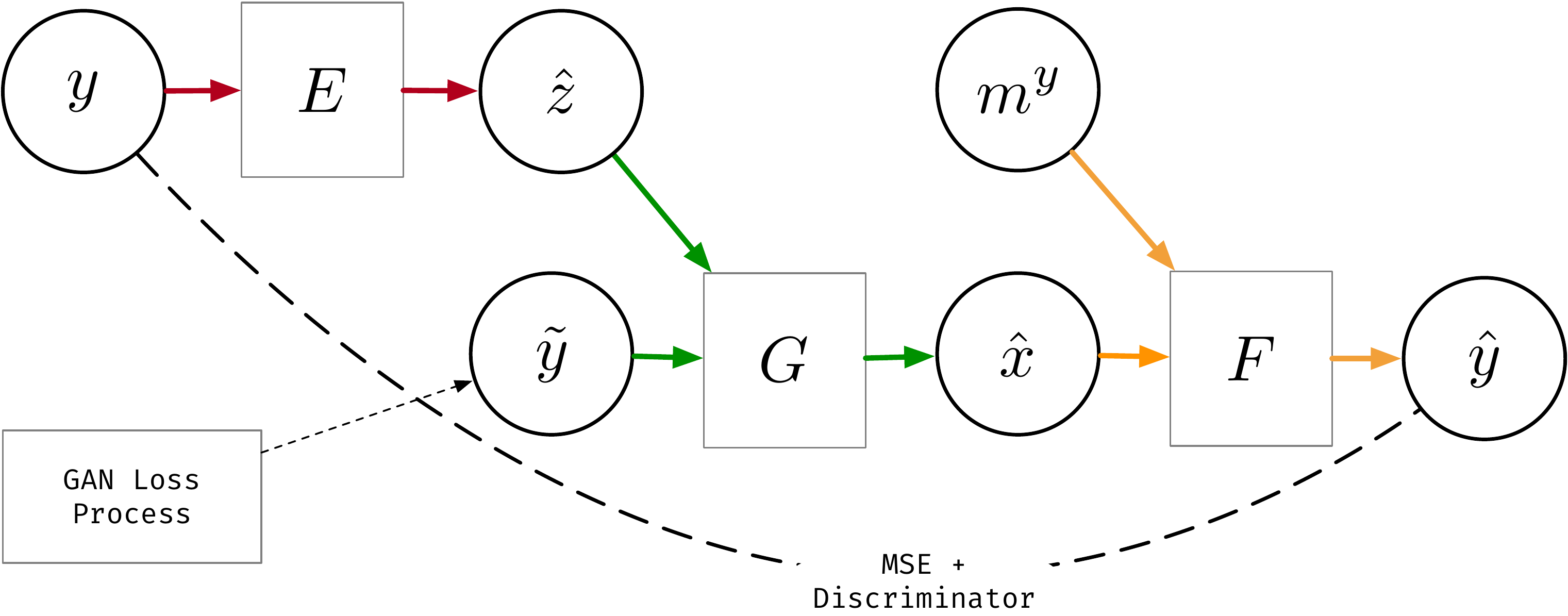}
\end{center}
  \caption{Encoding y reconstruction loss. the input masked images $y$ is encoded into the latent vector $\hat{z}$. The generator uses this code to generate a sample $\hat{x}$ which is masked with the mask $m^y$ shall produce $\hat{y}$ close to $y$.}
\label{fig:rec_z}
\end{figure}

An alternative to the above solution is to condition the $z$ value to a sample $y$. Let us start from the model in figure \ref{fig:gan_loss} and introduce a latent variable $\hat{z} = E(y)$ with $E$ being as before a mapping from $Y$ to $Z$. Let $\hat{z}$ and $\tilde{y}$, the latter being the output of the model in figure \ref{fig:gan_loss}, be mapped successively to $\hat{x}=G(\tilde{y},\hat{z}) $ and $\hat{y}=F(\hat{x})$ as illustrated in figure \ref{fig:rec_z}. Let us then constrain $\hat{y}$ to be close to $y$ via an Squared Error (MSE in figure \ref{fig:rec_z}), and let their distribution be similar via an adversarial loss (Discriminator in figure \ref{fig:rec_z})). Then $G$ will be forced to use both $\hat{z}$ and $\tilde{y}$. Let us briefly examine why. Thanks to the adversarial losses, $\hat{y}$ and $y$ will have the same distribution. However they are different since they are associated to different samples $m$. In order to have $\hat{y}$ close to $y$, the generator $G$ will then be forced to use $\hat{z}$. Using the above notations (see also figure \ref{fig:rec_z}), the corresponding auxiliary loss is:\\



\begin{equation}
\begin{split}
\mathcal{L}^{y}(G, D, E) & = \E_{\substack{y \sim \P_\ry}} \big[ \log D(y) \big] \\
 & + \E_{\substack{\tilde{y} \sim \P_\ry  \\ \hat{z} \sim{E(y)}}} \big[ \log\big(1 - D(F[G(\tilde{y}, \hat{z}) + \tilde{y}), m^y]\big) \big] \\
 & + \E_{\substack{y \sim \P_\ry \\ \hat{z} \sim E(y)}} \left\Vert y - F[G(\tilde{y}, \hat{z}) + \tilde{y}), m^y]\right\Vert_2^2
\end{split}
\label{eq:rec_y}
\end{equation}\\
In this equation, $\hat{z} \sim{E(y)}$ means that $\hat{z}$ is sampled from the distribution of $Z$ generated by sampling $E(y)$. $m^y$ denotes the  mask extracted from $y$ (see section \ref{sec:adversarial}). For simplification, we have omitted in the second and third expectation terms, samplings needed to generate $\Tilde{y}$.

This auxiliary loss is to be used in conjonction with loss in equation \ref{eq:adversarial_z_x}. The reconstructed image is, as before, $\Tilde{x}$.
\subsubsection{Putting it all together}
The two auxiliary losses (\ref{eq:rec_z})  and (\ref{eq:rec_y}) represent alternatives for enforcing the use of $z$. We have tested the two losses separately, each of them improves the quality and the diversity of the reconstructed images. we have found out that combining them improved further the image quality and diversity over individual solutions (see section \ref{sec:experiment}). This is coherent with the finding of \cite{zhu_toward_2017}. The complete objective is then a combination of losses (\ref{eq:adversarial_z_x}), (\ref{eq:rec_z}) and (\ref{eq:rec_y}): 

\begin{equation}
 \mathcal{L}(G, D, E) = \mathcal{L}^{GAN}(G, D)  + \lambda_{z} \mathcal{L}^{z}(G,E) + \lambda_{y} \mathcal{L}^{y}(G, D,E) 
\label{eq:loss_all}
\end{equation}

Where the values of $\lambda_{z}, \lambda_{y}$ are selected on a validation set. We describe the complete training process in the algorithm \ref{alg:deterministic}.

\begin{algorithm}[H]
\small
\setstretch{1.35}
    \begin{algorithmic}
        \Require 
        Initialize parameters of the generator $G$, the discriminator $D$ and the encoder $E$.
        \While{$(G, D, E)$ not converged}
            \State Sample $\{ y_i \}_{1 \le i \le n}$ from  data distribution $\P_\ry$
            \State Sample $\{ \tilde{m}_i \}_{1 \le i \le n}$ from $\P_m$
            \State Sample $\{ z_i \}_{1 \le i \le n}$ from $\P_\rz$
            \State Compute $m^y_i = T(y_i)$  
            
            \State Set  $\tilde{x}_i \leftarrow G(y_i, z_i) \odot \bar{m_i}  + y_i$ for $1 \le i \le n$
            \State Set $\tilde{y}_i \leftarrow F(\tilde{x}_i, \tilde{m}_i)$ for $1 \le i \le n$
            
            \State Set $\hat{z}_i \leftarrow E(\tilde{y}_i \odot \bar{m}^y)$ for $1 \le i \le n$
            \State Set $\tilde{z}_i \leftarrow E(y_i)$ for $1 \le i \le n$
            
            \State Set $\hat{x}_i \leftarrow G(\tilde{y}_i, \tilde{z_i}) \odot \hat{\tilde{m}}_i  + \tilde{y}_i$ for $1 \le i \le n$
            \State Set $\hat{y}_i \leftarrow F(\hat{x}_i, m^y_i)$ for $1 \le i \le n$

            \State Update $D$ by ascending: 
                \vspace{-.3cm}
                \State \begin{varwidth}[t]{\linewidth}\par
                \hskip\algorithmicindent $$\qquad \frac{1}{n}\sum_{i=1}^{n} \log D(y_i) + \log (1 - D(\tilde{y}_i))$$
                \end{varwidth}
            \vspace{.3cm}
            \State Update $G$ and $E$ by descending: 
                \State
                \begin{equation*} 
                \qquad\frac{1}{n}\sum_{i=1}^{n} \log (1 - D(\tilde{y}_i)) 
                \end{equation*}
                
                \State and 
                \begin{equation*}
                \qquad\frac{1}{n}\sum_{i=1}^{n} \lambda_{rec_y} \cdot \left\Vert \hat{y}_i - y_i\right\Vert_2^2 + \log (1 - D(\hat{y}_i)) 
                \end{equation*}
                
                \State and
                \begin{equation*}
                \qquad\frac{1}{n}\sum_{i=1}^{n} \lambda_{rec_z} \cdot \left\Vert \hat{z}_i - z_i\right\Vert_2^2  
                \end{equation*}
                
                \EndWhile
    \caption{Training Procedure.}
    \label{alg:deterministic}
    \end{algorithmic}
\end{algorithm}

\subsection{Implementation}

Our network architectures are inspired by the Self-Attention GAN architecture in \cite{sagan}.  They use residual networks, where each residual block of the generator and discriminator consists of 2 repeated sequences of batch normalization, ReLU activation, spectral normalization \cite{miyato2018spectral} and $3 \times 3$ convolutional layers. For the upsampling, we used a pixel shuffle layer \cite{shi2016real} for memory and speed improvement. The discriminator is similar to \cite{sagan}. For the reconstruction network G, we propose an image-to-image variant of their generator. The encoding network is a simple resnet with batch normalization and two fully connected network at the end, to predict the latent vector mean and variance.


When it comes to training the full network, the most sensitive parameters is the batch size. As demonstrated in \cite{biggan}, the learning benefit dramatically increases with scaling the batch size. The problem of the memory size of the network is alleviated by the gradient accumulation trick \cite{accumulation}.


\section{Experiments}
\label{sec:experiment}


\subsection{Datasets}
We made experiments on 3 datasets CelebA \cite{celeba}, Recipe \cite{recipe} and LSUN \cite{lsun} with similar conclusions. Due to a lack of place, we only illustrate the results on CelebA in the main text and provide additional examples in the supplementary material.

\vspace{-0.3cm}

\paragraph{CelebA} The CelebFace Attributes (CelebA) dataset contains $202599$ face images of celebrities. We  center crop $178 \times 218$  images to $128 \times 128$.
\vspace{-0.3cm}

\paragraph{Recipe} The Recipe1M  dataset is a dataset of cooked meals, containing  619424 samples. We resize the $256 \times 256$ images to $64 \times 64$.
\vspace{-0.3cm}

\paragraph{LSUN-Bedroom}  dataset of bedrooms, containing 3 million samples. We center crop the initial $218 \times 218$ size to $64\times 64$. 
\vspace{-0.3cm}

\paragraph{}
As we do not have any kind of supervision, we did not use a train/test split. However, in order to not overfit the data, we withhold $15 \%$ of the data , selected uniformly at random for each dataset in order to select the hyperparameters. To place ourselves in the most realistic setting possible, every image has been corrupted once, \textit{i.e.} there is never multiple occurrences of an image corrupted with different corruption parameters.

\subsection{Corruption}

We used two different masking processes. The first one denoted \textit{Patch(n,k)} consists in selecting $n$ patches of size $k \times k$ pixels. The top left position of each patch is uniformly sampled. These patches will correspond to an observation $y$, meaning that all the other non-selected pixels are set  to $0$. A small border of $4$ pixels in the image is also excluded in order to keep the background consistent. 
The second one denoted \textit{Drop Pixel} randomly samples a fraction $p$ of pixels uniformly and the three corresponding channel values are set to $0$.

\subsection{Training}

All models are trained using Adam \cite{kingma2013auto} with $\beta_1 = 0$ and $\beta_2 = 0.99$.  The batch size is set to $128$, for all experiments. We only update the networks every $4$ step, artificially setting the batch size to $512$. In order to train the generative networks, we use the non-saturating  adversarial hinge loss \cite{spectralnorm}. We also experimented with the minimax loss from \cite{gan} and the least square loss from \cite{mao2017least}, but obtained slightly more consistent results with the hinge loss \footnote{Code available at \href{url}{https://github.com/pajotarthur/unsupervised\_image\_inpaiting}}.

\subsection{Evaluation Metrics}
\label{sec:metric}

We used three complementary quantitative measures. As we are in an unsupervised setting, the evaluation metrics are distinct from the loss that we optimize.

\vspace{-0.3cm}

\paragraph{FID.} The Frechet Inception Distance \cite{heusel2017gans} samples reconstructed examples  $\tilde{x}$ and original uncorrupted examples $x$. The images are embedded into a feature space (the last layer of InceptionNet) where, the mean and  the covariance of the embedding are estimated. The Frechet distance between these two statistics is then computed. The score is an indicator of the visual quality of the generated samples.

\vspace{-0.3cm}

\paragraph{MSE.} We compute the Mean Square Error (MSE) between the reconstruction and the original uncorrupted image. This is not a significant measure of the visual quality, but it provides an indicator of the reconstruction capacity of the models. 

\vspace{-0.3cm}

\paragraph{Standard Deviation} To evaluate the diversity of the reconstructions, we compute the mean standard deviation of the reconstruction, over the mask reconstruted pixels, for $10$ different samples $z$, over $1000$ images. This correlates with the variation observed visually.

\subsection{Results.}
We first provide some examples of the images generated by the model for visual inspection and then detail quantitative results comparing our model to different baselines.

\subsubsection{Qualitative evaluation}

\begin{figure}[ht!]
\begin{center}
\includegraphics[width=\textwidth]{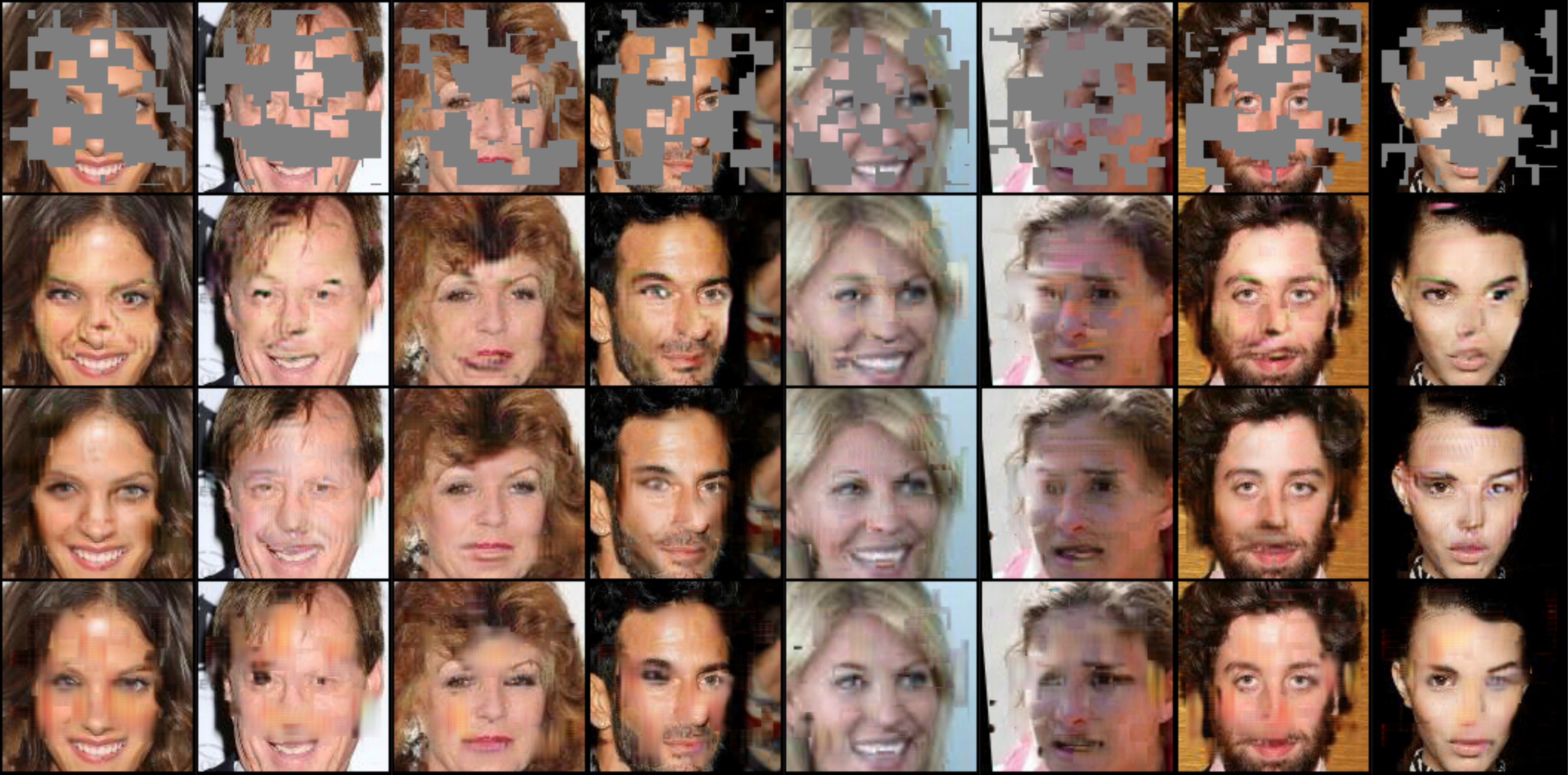}
\end{center}
   \caption{On the top row, randomly sampled measurements, from CelebA corrupted using \textit{Patch} with $n=90$ and $k=10$, and below associated reconstructions, for different latent vector $z$.}
\label{fig:result_ptit_trou}
\end{figure}

\begin{figure}[ht!]
\begin{center}
\includegraphics[width=\textwidth]{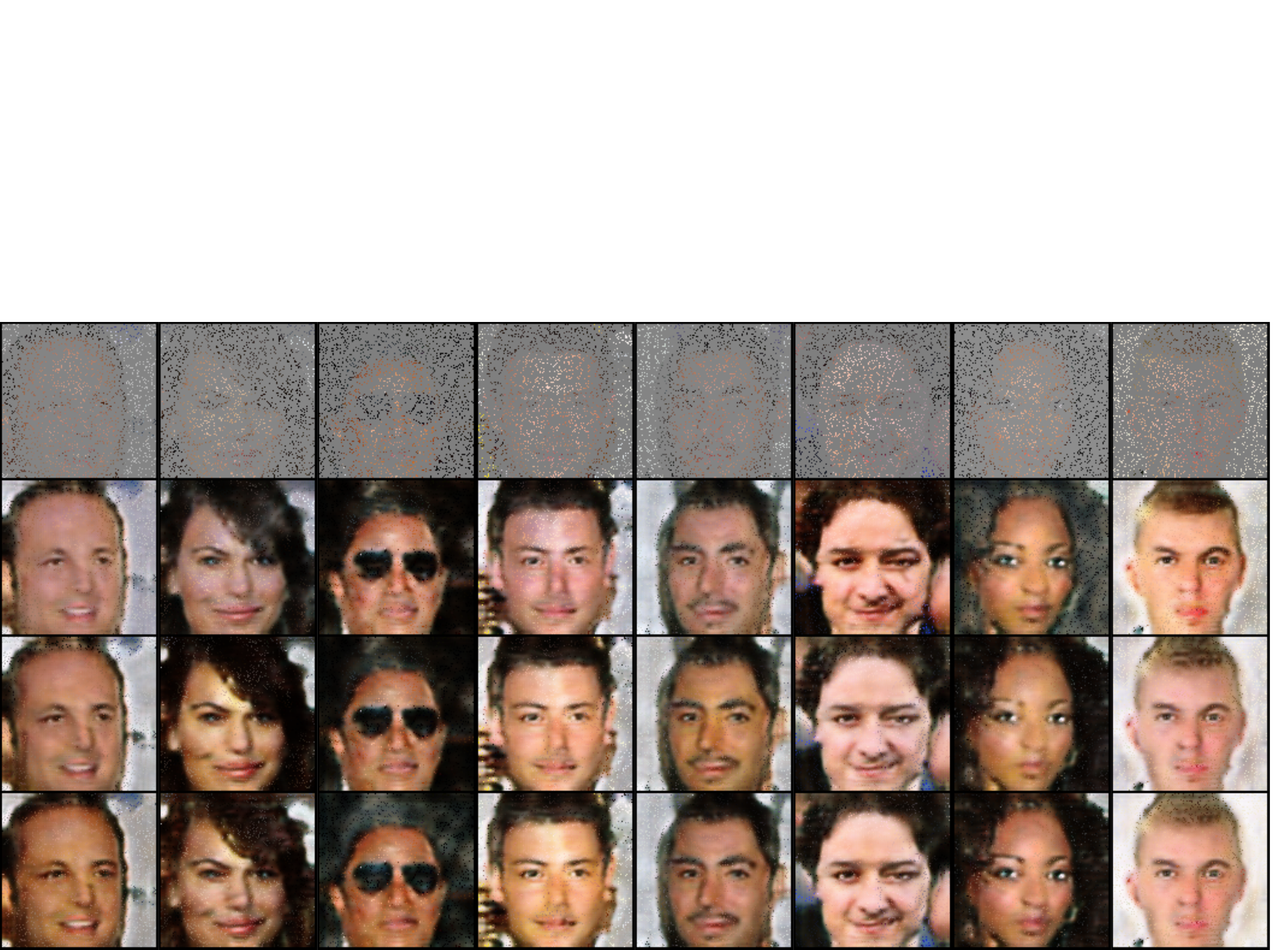}
\end{center}
   \caption{On the top row, randomly sampled measurements, from CelebA corrupted using \textit{DropPixel} with $p=0.90$, and below associated reconstructions, for different latent vector $z$.}
\label{fig:result}
\end{figure}

Figure \ref{fig:result_ptit_trou} and \ref{fig:result}  provide examples of reconstructions obtained  on the CelebA dataset with our model (equation \ref{eq:loss_all}), respectively for the \textit{Patch} and for the \textit{Drop Pixel} corruptions. The results are obtained using the model in figure \ref{fig:inference}. For each figure, the top row is the observation $y$ and the rows below are the associated reconstructions for different samples $z \sim \P_z$. For these experiments, the settings are respectively \textit{Patch(90,10}), i.e. 90 patches are selected from a full image $x$, each of size 10 pixels to build observation $y$, all the other values being set to 0, and \textit{Drop Pixel}, where $90 \%$ of the pixels are randomly selected and their values on the three chanels set to 0.

For each experiment, a specific training was performed for the selected corruption model, meaning that a training was performed for figure \ref{fig:result_ptit_trou} and another one for figure \ref{fig:result}. As can be seen, the reconstruction is not perfect, but given the large amount of corruption and the unsupervised setting, the model is able to reconstruct a large part of the information present in the original image, together with a significant diversity. The latter could be seen by inspecting the two figures \ref{fig:result_ptit_trou} and \ref{fig:result}. For example, the eyes most often come in different shapes and colours. Both figures exhibit random - not cherry-picked - samples. The reconstruction quality depends on the noise nature. It is easier for the \textit{DropPixel} corruption than for the \textit{Patch} one which is particularly difficult.
 

\subsubsection{Quantitative analysis}

\paragraph{}
\textbf{Ablation Study}
\paragraph{}

In order to analyze the importance of the different model components, we performed an ablation analysis, by training the model without any auxiliary loss (equation \ref{eq:adversarial_z_x}), with the \textit{Encoding z} auxiliary loss (equation \ref{eq:rec_z}), with the \textit{Encoding y} auxiliary loss (equation \ref{eq:rec_y}) and with both losses used together (equation \ref{eq:loss_all}). Results are reported in table \ref{table:ablation} for the CelebA $128 \times 128$ dataset for three corruption processes: \textit{Patch(1,32)} corresponds to a single patch of size $32 \times 32$ for the observation $y$, \textit{Patch(90,10)} corresponds to $90$ small patches of size $10$ pixels for $y$, \textit{Drop Pixel} corresponds to $90 \%$ of pixels selected with their three channel values set to 0. 
In table \ref{table:ablation} we report the values for the three quantitative measures: mean square error (MSE) scores between the reconstructed $\Tilde{x}$ and the true signal $x$ used to generate the input $y$, FID and the standard error deviation (see section \ref{sec:metric}). 

\begin{table}[ht!]
\small
    \caption{MSE, FID, and standard deviation on the celeb A dataset for three noises.}
    \label{table:ablation}
    \begin{center}
\begin{tabular}{c|c|c|c|c|c|c|c|c|c|c}
\multicolumn{2}{c|}{}       & \multicolumn{3}{c|}{One Patch}                    & \multicolumn{3}{c|}{Small Patch}                 & \multicolumn{3}{c}{Drop Pixel}                  \\ \hline
Encoding $z$ & Encoding $y$ & FID            & MSE            & std.            & FID            & MSE            & std.           & FID            & MSE            & std.           \\ \hline
0            & $\checkmark$ & 59.39          & 0.159          & 0.0463          & 20.37          & 0.062          & 0.024          & 69.73          & 0.130          & 0.058          \\ \hline
$\checkmark$ & 0            & 56.38          & 0.161          & 0.0408          & 26.39          & 0.062          & 0.024          & \textbf{63.07} & \textbf{0.089} & 0.074          \\ \hline
0            & 0            & \textbf{53.89} & \textbf{0.149} & 0.0325          & 28.42          & 0.069          & 0.014          & 74.82          & 0.131          & 0.026          \\ \hline
$\checkmark$ & $\checkmark$ & 54.55          & 0.156          & \textbf{0.0645} & \textbf{19.11} & \textbf{0.059} & \textbf{0.035} & 70.19          & 0.130          & \textbf{0.109}
\end{tabular}
    \end{center}
 \end{table}
 
 The three models implementing an auxiliary loss, all improve the diversity for the three corruption processes (\textit{std.} column) w.r.t. the simple base model (equation \ref{eq:adversarial_z_x}). The model combining the two auxiliary losses is clearly better than the others fort diversity. It thus seems better suited for learning to generate recinstcructed image distributions. For the large patch experiment, all the models exhibit very close performance, while the simp;le base model without any auxiliary loss performs slightly better. Here diversity comes at the expense of the reconstruction fidelity as measured by the FID and MSE criteria. For the multiple patches \textit{Patch(90,10)} experiments, the models with auxiliary loss all increase the performance w.r.t. the simple model ( equation \ref{eq:adversarial_z_x}), with the combined auxiliary loss clearly better than the other variants. For the \textit{Drop Pixel} experiment, again the proposed methods improve over the simple model, but here the \textit{Encoding z} version is better than the others.
 

\paragraph{}
\textbf{Comparison with baselines}
\paragraph{}

We also compared our model with several baselines. Note that in order to keep things comparable, all the networks used in the experiments are the same.
\paragraph{Unpaired Variant.} This is a supervised variant of our model where we do not have access to paired $(y,x)$ samples, but we have access to unpaired samples from $\p_\rx$ and $\p_\ry$. The baseline is similar to our model but instead of discriminating between a measurement from the data $y$ and a simulated measurement $\hat{y}$, we directly discriminate between samples $x$ from the signal distribution and the output of the reconstruction network $\Tilde{x}$. This should provide improved results w.r.t. the unsupervised model developed in the paper.

\paragraph{Paired Variant.} Here we have access to corrupted-uncorrupted pairs $(y, x)$ from the joint distribution $\p_{\ry, \rx}$. Given the masked image $y$, the reconstruction is obtained by regressing $y$ to the associated complete image $x$ using a MSE loss. In order to avoid blurry samples, we add an adversarial term in the objective, which helps $G$ to produce realistic samples, as in \cite{pix2pix}.

\paragraph{}For those supervised variant, we selected a $10 \%$ test set to evaluate our model.
\paragraph{MisGAN.} This baseline is adapted from \cite{misgan}. MisGAN (see the related work section) makes use of three generators;  one for learning the data distribution, one for the mask distribution and one for inputing the data, In our adaptation, since we suppose to  the mask distribution known, we replace the corresponding component in their model with the true corruption process as in our model.

\begin{table}[ht!]
    \caption{}
    \label{tab:supervised}
    \begin{center}
\begin{tabular}{c|c|c|c|c|c|c|c|c|c}
Model     & \multicolumn{3}{c|}{One Patch}                    & \multicolumn{3}{c|}{Small Patch}                 & \multicolumn{3}{c}{Drop Pixel}                  \\ \hline
          & FID            & MSE            & std.            & FID            & MSE            & std.           & FID            & MSE            & std.           \\ \hline
Unpaired  & 46.90          & 0.129          & 0.0359          & 18.55          & 0.053          & 0.015          & 60.28          & 0.098          & 0.032          \\ \hline
Paired    & \textbf{45.66} & \textbf{0.113} & -               & \textbf{18.32} & \textbf{0.044} & -              & \textbf{59.38} & \textbf{0.078} & -              \\ \hline
Misgan    & 84.63          & 0.166          & 0.0322          & 25.37          & 0.101          & 0.014          & 86.42          & 0.149          & 0.027          \\ \hline
Our Model & 54.55          & 0.156          & \textbf{0.0645} & 19.11          & 0.059          & \textbf{0.035} & 70.19          & 0.130          & \textbf{0.109}
\end{tabular}
    \end{center}
 \end{table}

As shown in table \ref{tab:supervised}, our unsupervised model (shown here with the combined auxiliary losses) reaches performance close to its variants trained using additional supervision. MisGAN shows slightly worse performance, due to the need to minimize the Generator and the Imputer GAN Losses. The convergence of this model is also order of magnitude slower. As before, the proposed model largely increases the diversity compared to the baselines meaning that it better learns a distribution of the reconstruted examples. MisGAN also makes use of a stochastic input as our model does but it is unable to generate examples with a significant diversity.




\section{Related Work}

\noindent \textbf{Inpaiting.}

Image completion and inpainting have a long history. Methods developed in the early 2000 relied either on diffusion, by optimizing an energy function, e.g.  \cite{Ballester2001,Bertalmio2000} or on patch completion, e.g. \cite{Simakov2008,Barnes2009}. There is no learning at all involved for these methods. Hole filling is usually performed by using only local information from the image itself and not attempting to extract information from other images.  There are some exceptions like \cite{Hays2007} who made use of image databases for completing the holes in the target image. 
More recently, convolutional neural networks were used for image completion. Contrarily to the present work, they typically assume some form of  supervision for mapping incomplete images onto reconstructed ones. Recent work usually relies on some form of adversarial training for obtaining sharp reconstructions.  A reference work here is  \cite{pathak2016context} which use content encoders in an image translation approach \cite{pix2pix} to map masked images onto non-masked ones. This is one of the first work demonstrating that large holes could be filled by a learning approach. \cite{multiscale_inpainting} extended content encoders by using a refinement network in which a blurry initial hole-filling result is used as the input and then iteratively improved. Yu and al. \cite{attention_inpainting}  use a post-processing step based on contextual attention layers. 
Non adversarial approaches have also been developed like e.g. \cite{partial_convolution} Liu and al. who introduce partial convolutions, where the convolution is masked and renormalized to be conditioned on valid pixel values only.

\noindent \textbf{Image to Image Translation.}

Inpainting can be considered as a special case of image to image translation for which several  GAN-based approaches have been proposed.
Recent works have achieved high quality results in image-to-image translation \cite{pix2pix,cyclegan,augmentedcyclegan}. For instance, pix2pix \cite{pix2pix} learns in a supervised manner by combining an adversarial loss with a L1 loss, thus requiring paired data samples. To alleviate the problem of obtaining data pairs, unpaired image-to-image translation frameworks \cite{cyclegan,augmentedcyclegan} have been proposed. Cycle GAN \cite{cyclegan} allows image translation by computing a bijection between the two domains. Augmented Cycle GAN \cite{augmentedcyclegan} and Bicycle GAN \cite{zhu_toward_2017} add stochasticity in the translation process by learning a mapping between a latent vector and the target domain. 
All these approaches then rely on some form of supervision either requiring paired or unpaired datasets. Unlike these image-to-image translation approaches, our model does not require access to images from the target domain. The work closest to ours is \cite{zhu_toward_2017} which makes use of similar auxiliary functions albeit in a supervised setting.


\noindent \textbf{Unsupervised Signal reconstruction.}

AmbientGAN  \cite{ambientgan} is an adversarial generative model aimed at generating images by being trained on noisy examples only and making use of a known stochastic measurement process as we did here. It does not perform completion but learns the distribution of the reconstruction space. Several works latter build on this model. \ 
Unsupervised Image Reconstruction \cite{pajot_unsupervised_2019} proposes a general method to solve inverse problems. The model is not specific to inpainting and can handle different types of corruption. On the other hand, it is not stochastic and does not attempt to learn a distribution. MisGAN
\cite{misgan} tackles the same problem as AmbientGAN \cite{ambientgan} in the specific case of masked images input like we do here. They extend the model in \cite{ambientgan} by assuming that the  mask distribution is unknown when it is supposed to be known for AmbientGAN. They train two generators, one as in \cite{ambientgan} for generating an image, and one for learning the mask distribution. The latter is trained as a classical GAN since the mask is fully observed. They propose a conditional variant of this model dedicated to the inputation task requiring the training of an additional generator. Compared to our approach, they also learn from incomplete data and do not rely on any supervision. Their hypothesis are even less restrictive than ours since they learn the nature of the noise itself. On the other hand, their model requires learning a generative model of the data prior to train a completion generator itself, when we address the problem more directly by training the inmputer on the incomplete observations themselves.
This clearly helps improve the performance by a significant margin.
\section{Conclusion}

We have proposed a new formulation for learning the distribution of inpainted images in an unsupervised setting, where only incomplete observations are available. The problem has been formulated using a Conditional Generative Adversarial Network setting. Training relies on the optimization of a criterion combining an adversarial loss allowing the recovery of sharp realistic signals, and a latent reconstruction loss allowing to condition the reconstructed image to an incomplete observation. Our experimental results show the relevance of this reconstruction approach. The unsupervised model is competitive with models that have access to higher forms of supervision.

\subsubsection*{Acknowledgments}
This work was partially funded by ANR project LOCUST - ANR-15-CE23-0027 and by CLEAR - Center for LEArning \& data Retrieval - joint lab. With Thales (\url{www.thalesgroup.com}).


%
\bibliographystyle{splncs04}
\bibliography{biblio}

\end{document}